\documentclass[conference]{IEEEtran}
\IEEEoverridecommandlockouts
\usepackage{cite}
\usepackage{amsmath,amssymb,amsfonts}
\usepackage{float}
\usepackage{algorithmic}
\usepackage{graphicx}
\usepackage{textcomp}
\usepackage{xcolor}
\def\BibTeX{{\rm B\kern-.05em{\sc i\kern-.025em b}\kern-.08em
    T\kern-.1667em\lower.7ex\hbox{E}\kern-.125emX}}
\begin{document}

\title{YOLO v3: Visual and Real-Time Object Detection Model for Smart Surveillance Systems(3s)\\

\author{\IEEEauthorblockN{Kanyifeechukwu Jane Oguine}
\IEEEauthorblockA{\textit{Department of Computer Science} \\
\textit{University of Abuja}\\
Abuja, Nigeria \\
jeaniettee@gmail.com}
\and
\IEEEauthorblockN{Ozioma Collins Oguine}
\IEEEauthorblockA{\textit{Department of Computer Science} \\
\textit{University of Abuja}\\
Abuja, Nigeria\\
oziomaoguine007@gmail.com}
\and
\IEEEauthorblockN{Hashim Ibrahim Bisallah}
\IEEEauthorblockA{\textit{Department of Computer Science} \\
\textit{University of Abuja}\\
Abuja, Nigeria\\
hashim.bisallah@uniabuja.edu.ng}
}
}

\maketitle

\begin{abstract}
Can we see it all? Do we know it All? These are questions thrown to human beings in our contemporary society to evaluate our tendency to solve problems. Recent studies have explored several models in object detection; however, most have failed to meet the demand in objectiveness and predictive accuracy, especially in developing and under-developed countries. Consequently, several global security threats have necessitated the development of efficient approaches to tackle these issues. This paper proposes an object detection model for cyber-physical systems known as Smart Surveillance Systems (3s). This research proposes a 2-phase approach, highlighting the advantages of YOLO v3 deep learning architecture in real-time and visual object detection. A transfer learning approach was implemented for this research to reduce training time and computing resources. The dataset utilized for training the model is the MS COCO dataset which contains 328,000 annotated image instances. Deep learning techniques such as Pre-processing, Data pipelining, and detection was implemented to improve efficiency. Compared to other novel research models, the proposed model’s results performed exceedingly well in detecting WILD objects in surveillance footages. An accuracy of 99.71\% was recorded with an improved mAP of 61.5.
\end{abstract}


\section{Introduction}
Deep Learning (DL), a notable buzzword in our contemporary society, has revolutionized the core implementations of Artificial Intelligence by advancing its scope of application in human endeavors. Fundamental reasons for its popularity and adoption include; architectural superiority in training machines to achieve human-centric tasks more accurately and efficiently, outstanding pattern recognition from relatively increasing amounts of data, and robust computational power (CPUs, GPUs, and TPUs). This paradigm has given rise to several research areas such as Computer Vision, Pattern Recognition, Natural language Processing (NLP), Biomedical Imaging, Genomic Engineering, Biometrics, Information Security, Human-Computer Interaction (HCI), and digital signaling, amongst others. Most research has employed deep learning to extract significant characteristics and solutions utilizing handmade features such as human crowd detection, emotion recognition, geo-location, monitoring, and identification.

Recent surges in global insecurity have propelled the need for several innovative research approaches to tackle them. A popular research area that has fast-tracked effective solutions to the issue of security and surveillance is Computer Vision and video surveillance systems. Despite broad adoption in public facilities like schools, hospitals, offices, airports, and private facilities, surveillance systems have seen an enormous shift from manual evaluation to intelligent processing given the relative amount of data generated in the 21st Century \cite{b15}; thus, \textbf {\emph{‘Smart Surveillance Systems’}}. Object identification over real-time video streams and visual image frames are currently used in Surveillance systems to detect variations and safety changes within the human environment, such as cases of theft, examination malpractice, robbery, fire outbreak, and weapon detection. However, gaps have been identified, showing the inefficiency of mundane evaluation  of footage and visual images from surveillance systems in making deductions. Hence, it necessitates the development of sophisticated, Intelligent, state-of-the-art research approaches to facilitate the detection, investigation, prevention, and deterrence of crime. \\
\textbf{Smart Surveillance Systems (3s):} Smart Surveillance systems (3s) are Cyber-physical Systems (CPS) proposed by this research that utilize an object Detection model based on the architecture of the YOLO v3 to help maximize the detection, Investigation, Prevention, and deterrence of security activities (Crime) in our society. 

Object detection can be referred to as the ability of computer systems to recognize instances of an object. Computer Vision, as highlighted earlier, is a Deep learning research area that has fostered the recurrent improvement of Object detection models, enabling the detection and classification of objects (animate and inanimate) in several interdisciplinary studies. This research field has enhanced several technological advancements in image processing solutions such as robotic vision systems, safety management software, object detection, and tracing. Significant reasons for these advancements are heralded by the need for efficient service delivery and unbiased recognition of behavioral patterns and environmental variations as typically observed from most Smart Video Surveillance Systems (VSS) implementing state-of-the-art models. Full-scale application instances and numerous security challenges case studied include UAVs, traffic monitoring systems, self-driving cars, Satellite and drone image feeds, monitoring sensors to detect and classify objects on visual images or real-time video frames, and the identification of vehicles on the road buildings and parking lots.
Deep learning-based object detection networks can be classified into two categories \cite{b5}:  \textbf {\emph{‘Single-Shot’ and ‘Region-Proposal’}} algorithms. The first category is the region-proposal algorithms such as (R-CNN, Fast R-CNN, Faster R-CNN, R-FCN, etc.) which make detections by systematically obtaining likelihood regions of concentration on input images before implementing classifiers (usually CNNs) in these regions. In spite of these models’ popularity for their high relative accuracies, slower than expected performance was observed because they process two rows on the image pixel matrix. However, Single-Shot algorithms like (SSD, RetinaNet, YOLO, etc.) have shown remarkable improvements in speed given robust frameworks that regress through the input image once to estimate the class and positions of the object using bounding boxes. This analogy has propelled a recurrent implementation of Single-shot algorithms in real-time detection compared to region-proposal. Despite the successes attained by earlier object detection models, such as You-Only-Look-Once (YOLO) \cite{b1}, Hybrid-DCNN model \cite{b2}, and Regional-based Convolutional Neural Networks (RCNN) \cite{b3}, \cite{b4} , in achieving high mean average precision (mAP) and accuracy; time complexities and frames per second (FPS) on non-GPU computers render them impractical for real-time use. 

This paper addresses the above mentioned problems by developing a deep learning model based on the YOLO v3 (You-Only-Look-Once, Version 3) framework for object detection in Smart Surveillance Systems (3s) and modifications put forward to improve the model’s efficiency and generalization. YOLO v3 is a Fully Convolutional Network model that employs bounding boxes and probability regression scores in making detections. It is one of the fastest real-time object detection algorithms in contemporary computer vision research solutions. The novelty of YOLO v3 as an object detection framework has broadened its myriad implementation in several research projects and solutions, primarily in object detection.  

\begin{figure}[htbp]
\centerline{\includegraphics[width=\columnwidth, height=15em]{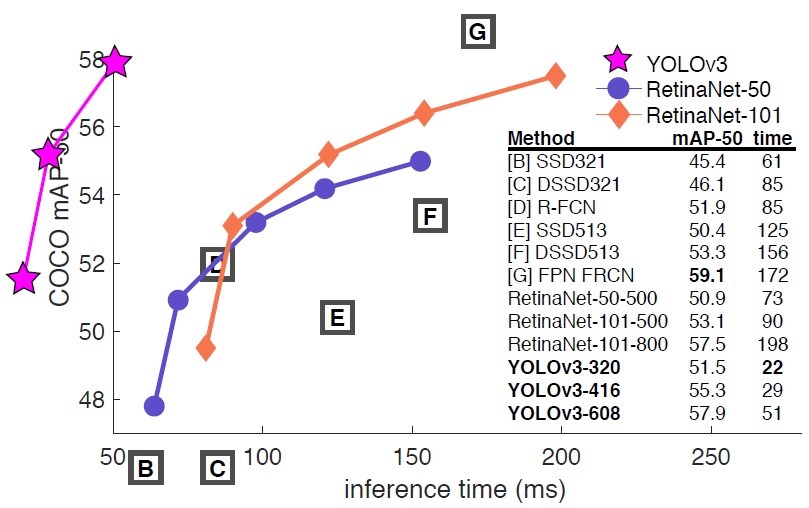}}
\caption{YOLO v3 vs. RetinaNet: Speed/Accuracy Chart \cite{b6}}
\label{1}
\end{figure}

Its impressive speed and accuracy in real-time object detection, when tested in comparison with RetinaNet using Pascal Titan X GPU on the MS COCO dataset (See Fig.1), emanates from the increased underlying architectural complexity commonly referred to as \textbf{\emph{Darknet}}. 
While preceding versions (YOLO and YOLO v2) struggled with drawbacks such as down-sampling of input images, absence of skip connections, and residual blocks, the Yolo v3 architecture was designed to contain features such as upsampling and residual skip connections. Another unique feature of the YOLO v3 is its ability to detect images at three different scales.

\section{Theoretical Background}
YOLO v3 framework was proposed in 2016 by \cite{b1} and has evolved to incorporate several advanced features paving the way for more accurate and precise state-of-the-art Object detection algorithms. Since its introduction, several versions (v4, v5, V6, and v7) of the YOLO framework have been developed, usually with increased depth through layer additions, with some primarily seen to have lightweight architecture, faster detection speed, and improvements in mean average precision (mAP). YOLO v3 algorithm implements Darknet-53 as a backbone for extracting features from input images. Darknet-53 is a deep neural network written in C and Compute Unified Device Architecture (CUDA) composed of convolutional layers used for feature extraction. It implements a Feature Pyramid Network (FPN) \cite{b7}, enabling feature maps extraction from input images. It supports CPU and GPU detection computations and is readily accessible on Github.

\begin{figure}[H]
\centerline{\includegraphics[width=0.7\columnwidth, height=17em]{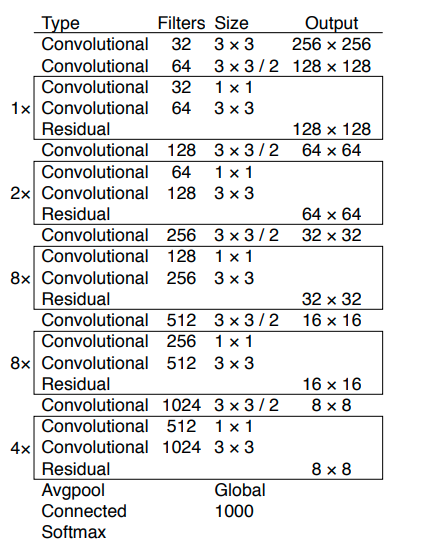}}
\caption{Description of Darknet-53 Framework.}
\label{2}
\end{figure}

The Darknet-53 network implemented on YOLO v3 architecture contains 53 layered convolutional neural networks followed by successive 3x3 and 1x1 fully convolutional layers that enable object detection (see Fig. 2). A further 53 additional layers were added, summing up to 106 layered convolutional structure, as shown in fig 2. YOLO v3 classifier is an ensemble of 106 fully convolutional layers. A novelty that makes the Yolo v3 more efficiently paced for real-time detection is its multi-scale prediction feature. Prediction made by YOLO v3 is made at three scales, which are precisely obtained by downsampling input image dimensions or strides by 32, 16, and 8, respectively (see Fig 3.).

\begin{figure*} %THIS CODE IS FULL SIZE IMAGES (SPANS 2 COLUMNS)
    \centering
    \includegraphics[width=6in]{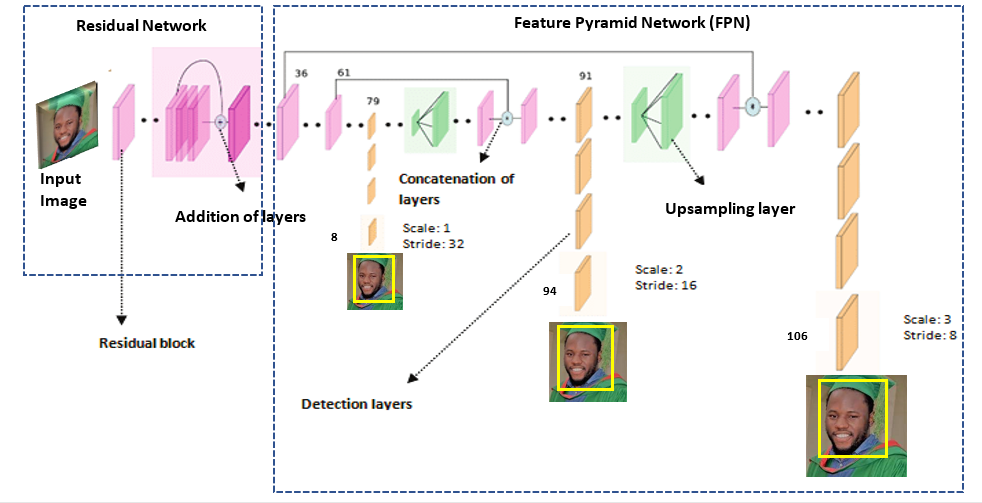}
    \caption{\textsf{YOLO v3 Architecture (Residual and FPN).}}
    \label{fig:ds}
\end{figure*}

Yolo v3 architecture utilizes the detection kernel $1 * 1 * (B * (5 + C))$ Where B stands for the relative number of bounding boxes a feature ma cell can predict, ‘5’ signifies four bounding box offsets and one object confidence, C is the number of probable classes. YOLO v3’s superiority has improved efficiency in detecting smaller objects (fine-grained features). An enormous reason for this is due to the concatenation of unsampled layers with fine-grained features (see Fig. 3). For multi-object detection from an input image, different layers such as $13 x 13$ responsible for the detection of large objects, $26 x 26$ for medium objects and $52 x 52$ for small objects are implemented for efficient object localization and detection in the YOLO v3 classifier. Three primary architectural updates in the YOLO v3 include:

\subsection{Residual Box and Skip Connections:}
 To avoid vanishing or exploding gradients, this research utilizes residual layers to determine the deviations from identity layers. This was necessary to prevent convergence degradation problems given the methodology of this research that requires outputs from one layer to be fed as input to the immediate or further layers in the neural network. Skip connections were utilized in this model instead of direct mapping (see Fig. 3) to ensure the direct transfer of input from one layer to another.
 
\subsection{Grid Cells:}
 An input image or real-time frame of a video is partitioned into grids with a dimension of S x S by the YOLO v3 classifier. These grids of equal dimensions (see Fig. 4) detect objects that appear within them. Residual blocks provide a chain known as Skip connections which ensures that inputs coming from the layer are fed into another layer. Below is a description of grids on an object image.

\begin{figure}[htbp]
\centerline{\includegraphics[width=\columnwidth,]{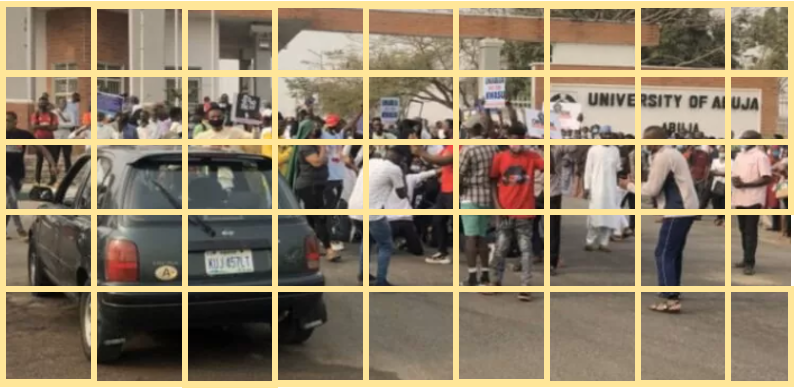}}
\caption{S x S Grid view of a Surveillance Video Frame}
\label{4}
\end{figure}

\subsection{Bounding-Box Regression:} 
Bounding boxes provide outlines that highlight objects in a visual image or video frame. YOLO v3 boasts more Bounding Box to Object ratio, improving object detection efficiency relative to the preceding architecture versions that were given by Eqn. (1). Output layers are used to predict dimensions, coordinates, confidence probability, and objectness score of a bounding box (see fig. 4). It implements anchors on the three prediction scales such that the entire architecture has nine (9) anchor boxes. Attributes of Bounding boxes are described below in $Eqn (1)$.
\begin{equation} \label{1} %THIS IS USED FOR EQUATIONS
	{x, y, w (width), h (height), Pr (confidence)}
\end{equation}
\textbf{Where}: \\
        Bounding box center $(bx, by)$\\
        Bounding box Width $(bw)$ \\
        Bounding box Height $(bh)$ \\
        Confidence Probability of class $(Pr (C))$ {such as person, car, traffic light, etc.}\\ 
\begin{figure}[htbp]
\centerline{\includegraphics[width=0.6\columnwidth, height=15em]{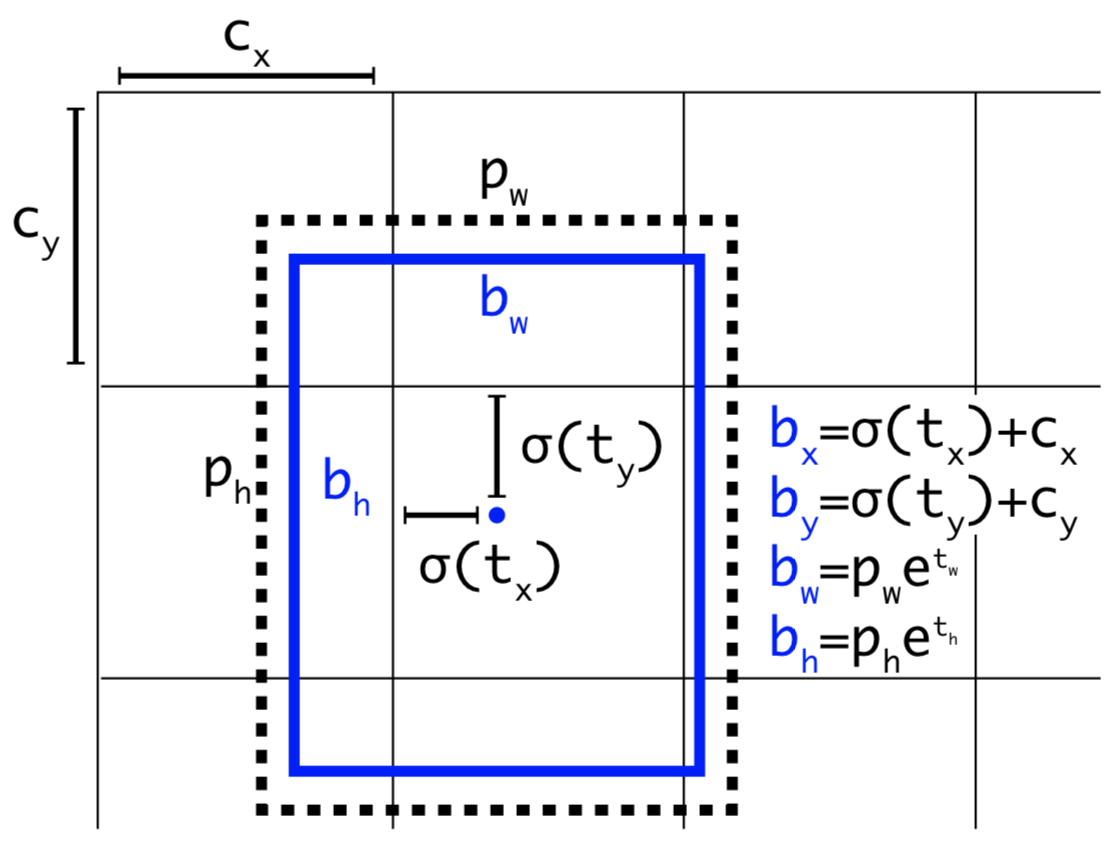}}
\caption{Transformations of Network Output Attributes of Anchors}
\label{5}
\end{figure} \\
In Fig. 5, values of network outputs $(tx, ty, tw, and th)$ are transformed into bounding box values (i.e., $bx, by, bw, bh$). $Cx$ and $Cy$ denote the top-left coordinates, while $Pw$ and $Ph$ represent the grids’ anchors.

\subsection{Intersection over Union (IoU):} 
IoU is an essential feature of the YOLO v3 classifier implemented by most state-of-the-art object detection models to describe the overlapping of bounding boxes. It provides a metric to evaluate the similarity between the predicting bounding box and the ground truth bounding box. The idea is to compare the ratio of overlapping area to the total combined region of the two boxes, as demonstrated in $Eqn (2)$
\begin{equation} \label{2} %THIS IS USED FOR EQUATIONS
IoU =	\frac{Overlapping Region}{Combined Region} 
\end{equation}
Object detection is done in the YOLO v3 classifier using bounding boxes and the principle of Intersection over Union (IoU). During object detection, a score of 1 denotes that the predicted bounding box precisely matches the ground truth box. A relative score of 0 implies that the predicted and ground truth boxes do not overlap. This mechanism is described in Fig. 6.
\begin{figure}[htbp]
\centerline{\includegraphics[width=0.9\columnwidth]{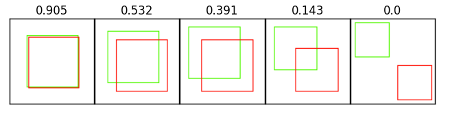}}
\caption{Sample IoU scores}
\label{6}
\end{figure}

\section{Proposed Methodology}
This research proposes an object detection model based on the YOLO v3 classifier for Smart Surveillance System (3s). Yolo v3 was utilized for this research because of its superior speed compared to other stable versions following recent updates. Compared with some newer versions \cite{b8}, Version 3 is a preferred choice that trades off a bit of accuracy for speed and is ideal for real-time detection in Smart Surveillance Systems (3s). Newer versions, despite their advantages, have some considerable drawbacks that this research takes into account. This research proposes a two-phased Smart Surveillance System (3s) model: the detection phase and the Analysis phase. These phases encompass pre-training and post-training processes, as described in Fig. 7 below.

\begin{figure}[htbp]
\centerline{\includegraphics[width=1\columnwidth]{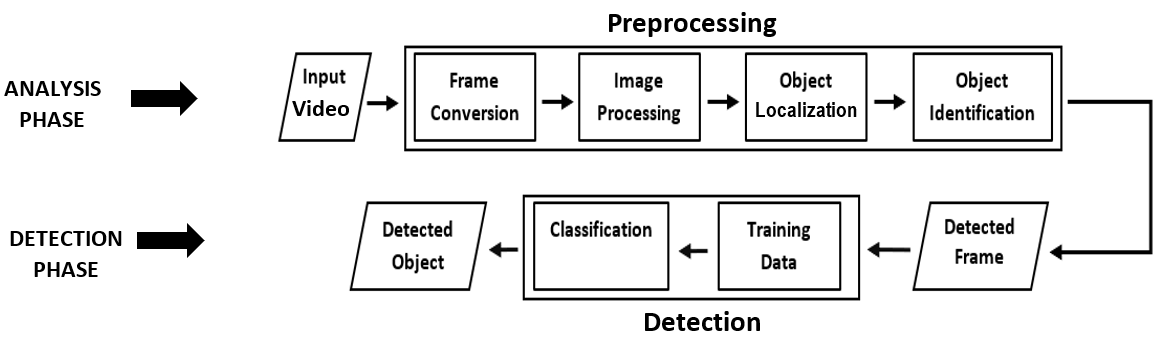}}
\caption{Flow Diagram for Research Methodology}
\label{7}
\end{figure}
\subsection{Analysis Phase:} 
The analysis phase of our proposed methodology consists of several pre-processing procedures, as shown in Fig. 7, which include the extraction of visual images or frames from real-time surveillance footage, pre-processing the image frames, and designing of the detection model and identifying localization of objects on a visual image or frame. The output of this phase is a pre-processed image frame which is then passed to the detection phase.

\subsection{Detection Phase:} 
The detection phase of our proposed methodology is very significant because it ensures the classification of detected objects. In this phase, the patterns in visual images or real-time frames detected from the analysis phase are passed in as input to the model. This process is followed by classifying the detected objects based on the categories highlighted. For this, we utilized a transfer learning approach where trained weights from the MS COCO dataset were used in setting up our model.

\section{Experimental Details}
\subsection{Dependencies:} For this study, a computing hardware specification (16GB RAM, Core i5 CPU, Jupyter notebook, and NVIDIA Quadro RTX 5000) was used. Libraries such as TensorFlow, NumPy, Pillow, Seaborn, and OpenCV v3.3.0 were utilized in setting up the development environment. Darknet-53 framework and CUDA 10.2 were also compiled using GPU computing to speed-up model training. Pretrained weights (yolov3.weights) from the Darknet codebase on the MS COCO dataset were imported.
\subsection{Dataset:} This research utilizes the MS COCO dataset for training the model. It is an extensive RGB image dataset with 328,000 images of everyday objects (bags, cars, bicycles, etc.) and living things (humans, dogs, cats, etc.) published by Microsoft. \\
\begin{figure*} %THIS CODE IS FULL SIZE IMAGES (SPANS 2 COLUMNS)
    \centering
    \includegraphics[width=6in]{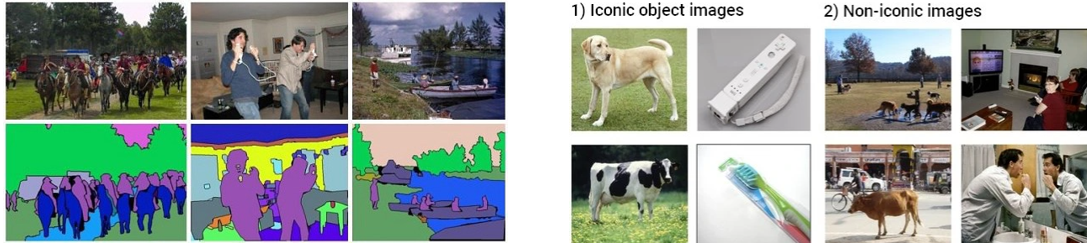}
    \caption{\textsf{(A):Image Classification Vs. Segmented Object Instances (B): Iconic and Non-iconic Images.}}
    \label{8}
\end{figure*}

\begin{figure*} %THIS CODE IS FULL SIZE IMAGES (SPANS 2 COLUMNS)
    \centering
    \includegraphics[width=6in]{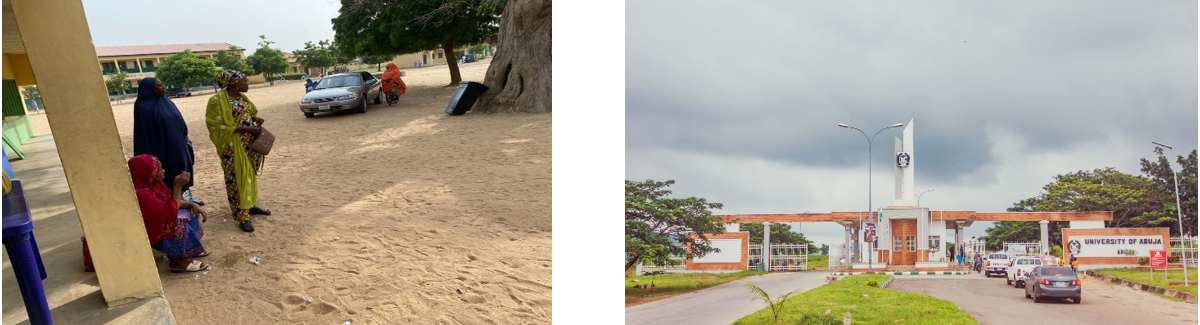}
    \caption{\textsf{(A): Frame from Surveillance Footage         (B): Aerial Visual Image taken by a Drone.}}
    \label{9}
\end{figure*}

The dataset provides annotation capabilities such as detection, captioning, keypoints, dense pose superpixelled stuff, and panoptic segmentation that improves a model’s recognition and description capability on up to 1.5 million object instances (see Fig. 8(A)). A name file (coco.names) which contains up to 80 classes was used for object detection and tracking. Model training based on the MS COCO dataset were either on iconic (simple) or non-iconic (complex) instances, as shown in Fig. 8 (B). For validation purposes, short surveillance footages from Nigerian universities and WILD non-iconic images taken by an E58 Folding UAV HD Camera selfie quadruple Helicopter drone were used to test the model (see Fig. 9(A) and 9(B)).
\subsection{Pre-processing:} This encompasses all the processes in the first phase of our model development. The transfer learning approach was used for improved performance since 90\% of the object’s class metadata to be detected in this research were already contained in the \textbf {\emph{coco.names}} name file. Also, a vital benefit of transfer learning in this research was to reduce training time and resources to the barest minimum. Annotation of video frames was not required, given the utilization of the transfer learning approach. Pre-trained weights \textbf {\emph{yolov3.weights}} obtained from training using the Darknet-53 code base were imported and stored in our working directory.

\subsection{Data Pipeline: } For this research, TensorFlow data APIs and Keras with optimized layers were defined to create a pipeline for data transmission and uploading model weights. A helper function was designed to create blocks of residual layers. Input test datasets for this research were resized, implementing three channels (RGB) and a target shape of $416 x 416$. The kernel size $(K_{s})$ is given by $1 * 1 * 255$ in $Eqn. (3)$.

\begin{equation} \label{3} %THIS IS USED FOR EQUATIONS
\begin{split}
K_{s} =	1 * 1 * [B * ((4 + 1) + C)] \\
K_{s} =	1 * 1 * [3 * (5 + 80)] \\
K_{s} = 1 * 1 [3 * (85)] \\
K_{s} = 1 * 1 * 255
\end{split}
\end{equation}

The $Eqn (3)$ above specifies the mathematical computation of the detection kernel size utilized by this model. Here B represents the number of bounding boxes predicted at each scale, (4 + 1) describes the bounding box offsets plus one object confidence, and C represents the number of class predictions.
\subsection{Upsampling:} When an image is passed through several layers in a network, the image quality degrades, causing the omission of several features. To prevent this, our research upsamples each image at various layers in the network.

\subsection{Non-maximal Suppression:} In ensuring precision and detection accuracy, a threshold probability of 60\% was defined to ensure that only the confidence score of objects higher than 0.6 is outputted. During prediction, it was observed that the model predicted a lot of individual bounding boxes for object instances on the input image frame. This experiment employs IoU and bounding box filters referred to as non-maximal suppression to overlap the bounding boxes and reduce them to a single box with \\
\begin{figure*} %THIS CODE IS FULL SIZE IMAGES (SPANS 2 COLUMNS)
    \centering
    \includegraphics[width=6in]{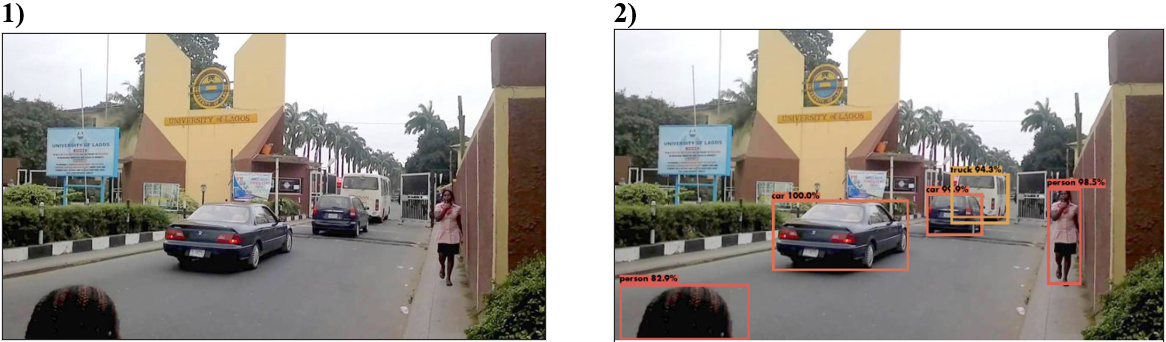}
    \caption{\textsf{(A):Image Classification Vs. Segmented Object Instances (B): Iconic and Non-iconic Images.}}
    \label{10}
\end{figure*}

\begin{figure*} %THIS CODE IS FULL SIZE IMAGES (SPANS 2 COLUMNS)
    \centering
    \includegraphics[width=6in]{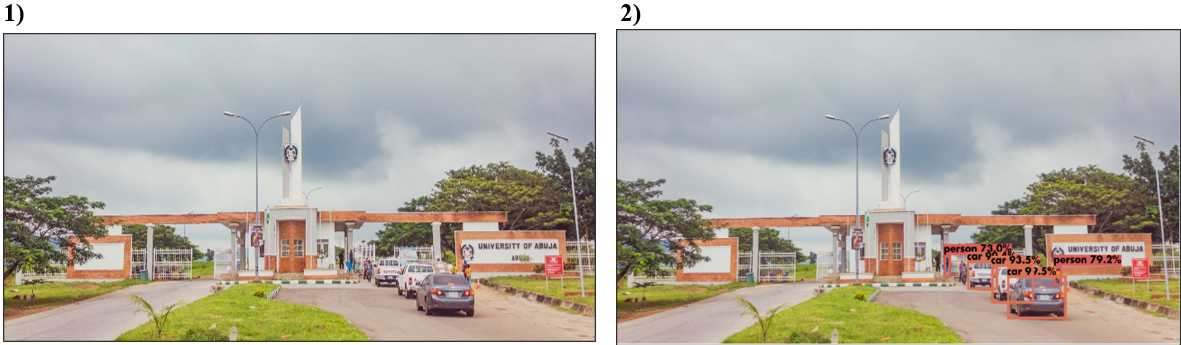}
    \caption{\textsf{(A): Frame from Surveillance Footage         (B): Aerial Visual Image taken by a Drone.}}
    \label{11}
\end{figure*}
the configuration parameter of 0.6. Classification of predicted objects was achieved by enumerating all boxes and comparing their various class prediction values. Corresponding string class labels from the name file in the working directory were assigned to detected objects see fig. 10 and 11.

\subsection{Detection and Prediction:} To make predictions using our model, test datasets are required, including new footage frames and visual images, as shown in Fig. 10 and 11 above. The test datasets (zeeDATASET) are uploaded to the working directory. Output images are predicted within bounding boxes using three different grid sizes. However, to visualize the bounding boxes on an output image, the shape of test footage frames or visual images was changed to (m, 4), where m denotes the number of bounding boxes with high confidence scores.

Transformation of bounding boxes coordinates from the YOLO format to the coco format was done using these steps, namely:
\begin{itemize}
    \item Return the scale of the boxes from 0-1 to 0-256
    \item Transform the x, y plane of the box from the centre to the top left corner
    \item Convert width and height to $x_max, y_max$
\end{itemize}

\section{Results}
For this research, Average Precision (AP) was used as a measure to evaluate the performance of our model. The results obtained from each detection are compared with the ground truth before they can be counted as True positive (TP). Hence, the IoU confidence score of the bounding boxes and corresponding ground truth boxes must be greater than 60 \% (IoU $>$ 60). To obtain the accuracy of our model quantitatively, the following mathematical computations are in Eqn. $(4,5,6)$ were utilized:

\begin{equation} \label{3} % EQUATION FOR ACCURACY
    Accuracy = \frac{TP + TN}{TP + TN + FP + FN}  
\end{equation}
\begin{equation} \label{3} % EQUATION FOR PRECISION
    Precision = \frac{TP }{TP + FP}  
\end{equation}
\begin{equation} \label{3} % EQUATION FOR RECALL
    Recall = \frac{TP}{TP + FN}  
\end{equation}
\textbf{Where}: \\
\textbf{TP:} Positive object class is predicted correctly; \\
\textbf{TN:} Negative object class is predicted correctly; \\
\textbf{FP:} Positive object class is incorrectly predicted; \\
\textbf{FN:} Negative object class is incorrectly predicted. \\

In instances of multiple detections of the same object, only a single detection can be counted as a True Positive. This research considered three (3) separate videos to validate and test the model’s generalization capability. This validation was based on three benchmarks (See Table 1). The video sequence is broken into frames to get the mAP value during real-time object detection from surveillance footages. Tracking information of each frame in a video was extracted and saved.

\begin{figure*} %THIS CODE IS FULL SIZE IMAGES (SPANS 2 COLUMNS)
    \centering
    \includegraphics[width=6in]{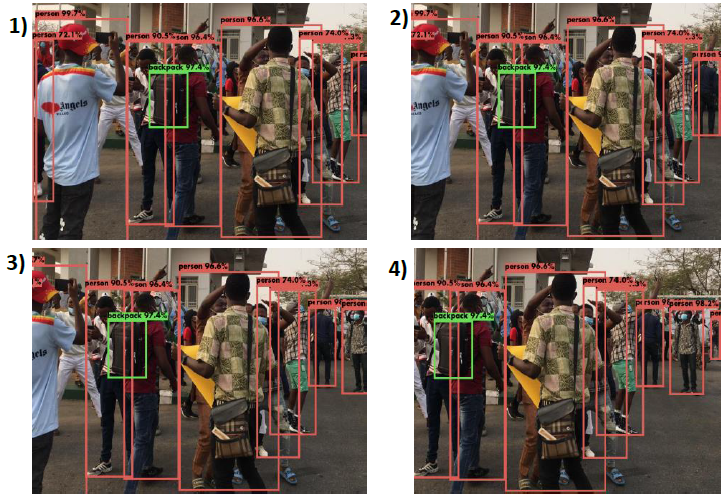}
    \caption{\textsf{Annotation of objects on Video Frames from surveillance footage..}}
    \label{12}
\end{figure*}

\begin{table}
\caption{Comparative Evaluation Metrics for Test Videos}
\label{1}
\begin{tabular}{|*{12}{p{1.32cm}|}}
\hline
\textbf{Video}  & \textbf{Frame Numbers} & \textbf{Accuracy} & \textbf{Precision} & \textbf{Recall}  \\ \hline
1  & 543    & 0.753 & 0.928     & 0.924    \\ \hline
2  & 621    & 0.815 & 0.896     & 0.997     \\ \hline
3  & 412    & 0.701 & 0.874     & 0.923    \\ \hline
4  & 765    & 0.892 & 0.944     & 0.956     \\ \hline
\end{tabular}
\end{table} 
From Table 1 above, three videos were analyzed comparatively based on three (3) benchmarks (accuracy, precision, and recall). Fig. \ref{12} shows the output frames of the test video. 

\begin{table}
\caption{Comparison with Existing Research}
\label{2}
\begin{tabular}{|*{12}{p{1.70cm}|}}
\hline
\textbf{S/No.}  & \textbf{Models}  & \textbf{Dataset} & \textbf{Accuracy} \\ \hline
1 & YOLOv3 + YOLOv3-tiny \cite{b9}  & MS COCO + VOC dataset   &  90\%    \\ \hline
2 & CNN VGG-16 \cite{b10}       & IMDB                        & $>$ 90\%     \\ \hline
3 & Alexnet + SVM \cite{b11}    & Gun video database          & 95\%    \\ \hline
4 & Faster RCNN \cite{b12}      & Streaming video             & 95.4\%     \\ \hline
5 & Custom YOLO v3 \cite{b13}   & Private Image dataset       & 98.89\%     \\ \hline
    \textbf{6} & \textbf{Our trained model YOLO v3}     & MS COCO + Video dataset collected & \textbf{99.71\%}    \\ \hline
\end{tabular}
\end{table}
Table 2 comparatively describes models developed by other researchers and the dataset employed in testing and training the model. As can be observed from the table, this research utilizes the MS COCO dataset and Video footages collected in WILD environments. As can be observed from the table, our model has a detection accuracy of 99.71%, in contrast to the accuracy of preceding research.

\section{Conclusion and Future Work}
A major challenge in most surveillance systems today, especially in developing and underdeveloped countries, has been the inadequacy and inefficiency of handling proper object detection. Traditional/human surveillance techniques in security surveillance have proved useless in tackling the ever-increasing security threats. This paper proposes a new video surveillance approach that should be employed for effectively tracking and detecting objects. Smart Surveillance System (3s) entails the application of Deep learning approaches to surveillance. The impact of this approach is to foster the creation of Intelligent, cyber-physical systems for object detection with great precision and accuracy. YOLO v3, a novel deep learning architecture for object detection, was implemented in designing the framework of our model. This architecture was utilized for this research because of its wide adoption and speed in detecting objects, which are vital features of surveillance systems. Holistic emphasis was made in describing the superior theoretical background of the YOLO v3 Architecture. The multi-scale capability, as discussed in this research, was an essential factor in its widespread utilization including in this research. Features such as residual blocks, skip connections, upsampling and grid splits were also discussed and practically explained in the experimental phase of this research. For real-time detection of objects, localization coordinates of objects on extracted video frames were used to generate detection bounding boxes. The principles behind the object detection mechanism were explained and can be seen from the experimental results in this research. Overall, this research will significantly improve video surveillance in areas where security is paramount and lesser computing resources are available. This research recommends that other sophisticated architectures be employed in Security Surveillance for future studies. Also, the cyber-physical system used in this research be implemented. Customized neural network architectures are utilized to streamline surveillance to subdomain fields.

\section{Abbreviations}
YOLO – You-Only-Look-Once \\
NLP - Natural Language Processing\\
DL - Deep Learning \\
HCI - Human-Computer Interaction \\
CPU - Central Processing Unit \\
TPU - Tensor Processing Unit \\
GPU - Graphical Processing Unit \\ 
VSS - Video Surveillance Systems \\
UAV - Unmanned Aerial Vehicle \\ 
DCNN - Deep Convolutional Neural Network \\
FPS - Frames per Second \\
mAP - Mean Average Precision \\
IoU - Intersection over Union \\
FPN - Feature Pyramid Network \\
CPS - Cyber-physical systems \\
CUDA - Compute Unified Device Architecture \\
SVM - Support Vector Machine \\
MS COCO - Microsoft Common Objects in Context 

\section{Declarations}
\subsection{Disclosure}
The authors declare that they have no competing interests.

\subsection{Author Contributions}
All authors contributed significantly to conception and design, data acquisition, analysis, and interpretation; participated in compiling the article and critically revising it for significant intellectual content; agreed to the submission.

\subsection{Data Availability Statement}
This research paper utilizes the MS COCO dataset for model training [13]. Real-time surveillance footages were also obtained and granted permission to be utilized for model testing. \\

\vspace{12pt}
\end{document}